# Introduction to the iDian

## A natural language platform for computer operation

**Alan (Xin Rong)**

Team: Xin Rong, Chenguang Ma, Lin Lin,
Xuan Zhang, Surong Ruan, Xiaochun Yi
Dept. of Automation, Tsinghua University
**alanrong89@gmail.com**
http://sites.google.com/site/alanrongxin/

## Abstract

The iDian (previously named as the Operation Agent System) is a framework designed to enable computer users to operate software in natural language. Distinct from current speech-recognition systems, our solution supports format-free combinations of orders, and is open to both developers and customers. We used a multi-layer structure to build the entire framework, approached rule-based natural language processing, and implemented demos narrowing down to Windows, text-editing and a few other applications. This essay will firstly give an overview of the entire system, and then scrutinize the functions and structure of the system, and finally discuss the prospective development, esp. on-line interaction functions.

**Note:** The iDian is designed and implemented in simplified Chinese environment, and the syntactic and semantic models only accept Chinese as input. Any command that is underlined in this document is a translated version, which won't be recognized by current iDian's demos in its English form.

## Background

Current human-computer interactions rely on graphical user interface. The costs and difficulties of GUI designing increases as the software applications getting more complicated and integrated, nonetheless the users find it more difficult to find expected orders efficiently.

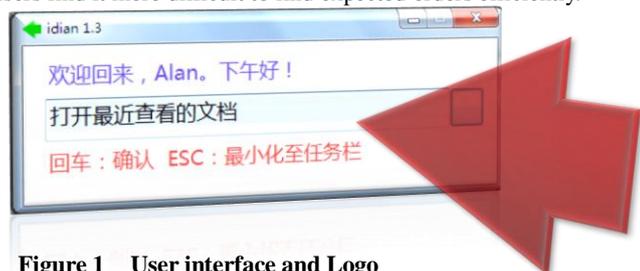

**Figure 1    User interface and Logo**
The Chinese sentences say: (1) iDian v1.3; (2) State: Welcome back, Alan. Good afternoon! (3) Open the most recent document (4) Enter: Confirm ESC: Hide

The iDian is an interface between users and applications, making it possible for users to input actively and get what they need, instead of passively searching lists of options. This way of interaction is much like a command-line interface; nevertheless the entire framework is fundamentally different, providing exciting functions and user experience. For example, if the user types "Delete carriage returns in each line", iDian will make *Word* perform this promptly, otherwise the user would have to delete them line by line, or type regular expression in the Replace Dialog, like "^p", which is not accessible to all people. Moreover, if the user inputs "I want to make a video call with my son", iDian will firstly ensures the online-communication software to be used, and then searches the contact-list and make a call automatically, which is especially useful for computer beginners or "permanent intermediate users".

The iDian provides an entire interaction solution, rather than a front user interface. By providing the special suit, software developers could easily add support for iDian to their programs. Therefore, the iDian builds up a bridge between the manufacturer and the customer of software, boosting up the way people interact with personal computers and associated devices.

## Functions

In order to make the readers get a clearer view on iDian's functions, here we present several more detailed situations where iDian is used to increase efficiency and help computer beginners.

**Example 1:** Dealing with formats in *Word*.

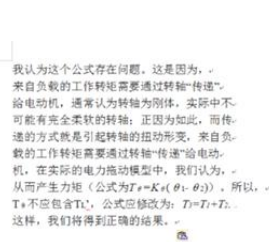

**Figure 2**
Redundant carriage returns in a document copied from a webpage.

*Microsoft Word* is very "convenient" to work with in most cases. However, it could be smarter if more complex orders are accepted. The iDian enables users to manipulate *Word* in format-free expressions which are exactly natural language, so that with only one typing they could complete a task that might have otherwise taken several steps of selections and clicks. Here is a direct example:

A passage copied from a webpage (or PDF document) to *Word* is usually filled with redundant carriage returns (as in Figure 2), which decrease the efficiency when the document is being edited. The solution of most users dealing with similar problems is to delete the carriage returns line by line, which is slow and obviously violates *Word*'s principles. The "correct" way is: first select the paragraphs, and then press Ctrl+H to call up the Replace Dialog, and then input expressions as below:

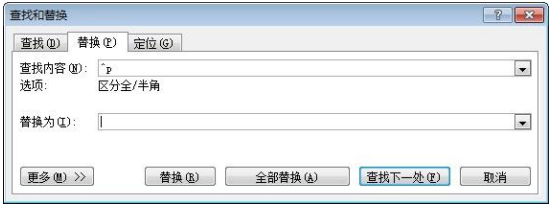

**Figure 3** The "correct" way to remove redundant carriage returns is to use regular expressions.

In iDian's way, all these steps are reduced into typing one sentence that directly describes the requirement: "<u>Delete carriage returns in each line</u>". By typing this sentence, the user need not to care about the implementation within iDian, and just need to wait up a few milliseconds to see the task completed as in Figure 4. In fact, lots of operations could be performed in this way, all of which require only one sentence as input, like "<u>Transform numbers in lines 1-3 to inferior characters</u>", and "<u>Email this document to *Xi Wang*</u>", etc. Some of the operations appear to be a piece of cake for skilled computer users, but some are not. For computer beginners and the numerous "permanent intermediate users", the reason why they prevent using advanced functions of a program is not that they have none such requirements, but that they are tired or afraid of searching the menus. By using iDian, the obstacles are overcome for people to approach advanced functions and combinations of basic functions of applications.

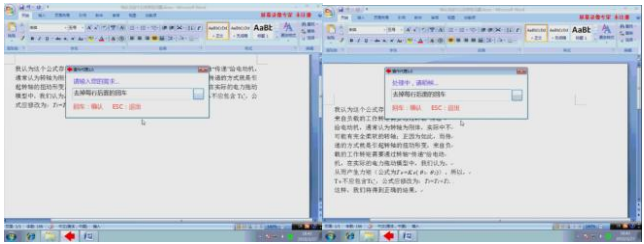

**Figure 4** Remove the redundant carriage returns using one sentence.

**Example 2:** Learning and operating *Maya*.

Here is another example: *Autodesk Maya* is a famous software application used for 3D-modelling and animation. Although it's said that *Maya* is easy to learn, it is really difficult to master. Through iDian, *Maya* beginners are easily to get access to basic functions and get familiar with the operations more quickly, while advanced users will save much time for basic operations.

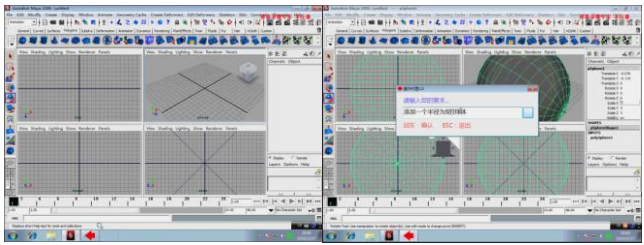

**Figure 5** Use iDian in Maya to "Create a sphere with a 5 radius"

Sometimes the user wants to see quick effects of an operation. In such cases, using iDian becomes the best solution, where there is no need to adjust various parameters in dialogs; everything the user has to do is just to state her/his demand and press Enter before she/he abandons trying the idea. It does like command-line to some extent, yet it provides much freer space for users to organize their expressions much more naturally.

The iDian's support for *Maya* is currently experimental, which means the amount of available actions is limited at the moment. We used MEL language as the interface to control *Maya*, and are planning to provide online-help functions.

## Structure and Mechanism of iDian

The iDian's employs a multi-layer structure, consisting of a syntactic layer, a semantic layer, an explainer, an executor, and a leaner. Figure 6 depicts an outline of the structure and workflow within iDian.

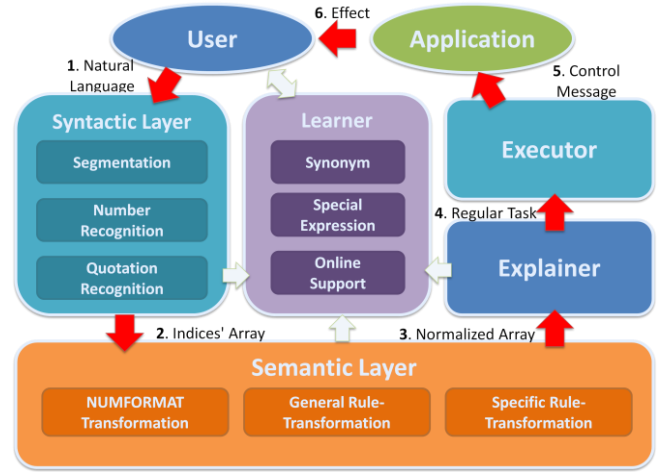

**Figure 6** The structure and workflow inside iDian.

**Syntactic Layer:**

As demonstrated by Figure 6, when the user types a sentence in natural language, the sentence is firstly imported into Syntactic Layer, where the sentence is segmented, looked up in dictionary and transformed to integer numbers. Due to the synonym phenomenon, there is usually more than one word corresponding to an index number. The quotations are also extracted at this step, where a temporary index will be distributed to each of the numbers and quotations.

**Semantic Layer:**

Then the indices' array comes to Semantic Layer with quotations appended. In this layer, a series of number-format transformations are firstly performed, resulting in NUMFORMAT, a structure that describes all types of numbers, including **ordinal numbers**, **cardinal numbers**, **ranges** and **arrays**, and an additional flag is set to distinguish **absolute numbers** and **relative numbers**.

After NUMFORMAT transformation, general rule-based translation and specific rule-based translation are separately performed. The distinction between the two kinds of transformations is that: the former is suitable to all applications, and the latter is specific to each application. Several special **wildcard characters** are employed, including "*N", "?N", "#N" and "!N", where N corresponds to word ordering or class serial number, "#" stands for temporary classes, and "!" stands for part-of-speech classes. We found that transformations with these four kinds of wildcard characters are sufficient to transform ordinary imperative sentences in Chinese into regular expressions that are recognizable by Explainer in the next step. Although the models are carefully written so that the user's commands could be correctly parsed as many as possible, recursive or hierarchical combinations of modifiers in the sentence are not well supported by current rule-based model – the semantic layer will sometimes produce incorrect results by making wrong assumptions. This problem should be resolved by introducing more knowledge about semantic processing or employing statistical methods based on corpus and large datasets.

**Explainer:**

The output of Semantic Layer is still a number array, but the whole sentence is rearranged into regular order, so that Explainer in the next step could directly separate the parts of the sentence. The functional parts of the sentences are tagged **Action** (VP), **Primary Object** (NP), **Secondary Object** (NP, Optional) and **Conditions** (PP, Optional), and then delivered to Executor whenever a clear end or start of a tagged-part is recognized. An "End" message is sent by Explainer when the input array reaches its end, and will start the execution process in the final step. For example, if the input sentence is "Replace 'apple' with 'peach' in lines 1-10 that contain 'orange' and 'bread'", then the output of semantic layer will be like "Replace in [Number 0] line has-and [Quotation 2] [Quotation 3] [Quotation 0] with [Quotation 1]" (the words are represented by numbers in reality). And Explainer will tag the elements of the command as follows:

| Tag | Content | Index | Corresponding Extra Data |
|---|---|---|---|
| Action | Replace | **1011** | |
| Primary Object | [Quotation 0] | 5000 | 'apple' |
| Secondary Object | [Quotation 1] | 5001 | 'peach' |
| Condition [0] | In [Number 0] line | **3002** 6000 **2015** | [Absolute][Range] 1, 10 |
| Condition [1] | Has-and [Quotation 2] [Quotation 3] | **3005** 5002 5003 | 'orange' 'bread' |

**Table 1** **Example output of Explainer.** Note: All language models are actually based on Chinese, while this is only a translated version.

In the example above, the preposition "with" is recognized as a switch between primary object and secondary object. Besides, there are three quotations appear continuously in Semantic Layer's output, but only Quotation 2 and 3 are recognized as part of Condition 1, due to the regulation of the previous preposition "has-and", which is produced during the general rule-based transformations.

**Executor:**

Upon receiving the "End" message from Explainer, Executor begins to perform the operation by sending messages to application. Implementations of Executor varies with the method it controls the target application. Although it is quite complicated to realize Executor, it has much less theoretical difficulties than the previous steps. Also, we are expecting newly distributed programs to actively offer a better interface for iDian.

Usually, Executor performs the operation in a similar way that IMEs (Input Method Editors) send messages to applications. However, Executor will not make one-one mapping between the combinations of regular tasks and the exact functions of the software. After all, the number of the combinations is infinite, therefore inside Executor a mechanism is established to verify all the conditions and map the primary object and secondary object to real targets in the application. This process ensures that every reasonable expression generated by the user would be precisely performed in its original intention, as long as it is "feasible" by **direct combinations** of currently supported operations, where "direct" means that every element of the expression has to be conceptually clear and well-defined whether in the application or within Executor. For example, the expression "Make an outline of the last two paragraphs" will possibly get through the first three layers, but will confuse Executor, unless "outline" as a primary object under the action "make" is mapped to the function "Generating Abstract" of the application, or defined as a process within Executor.

The user interface of iDian will not permanently stay on top of the screen – its behavior is much like a "**transient posture program**", which "is invoked when needed, appears, performs its job, and then quickly leaves, letting the user continue her normal activity, usually with a sovereign application." [Alan Cooper, *About Face 3*] Only when the user is confused with some specific operation of a program, she/he will "wake up" iDian for help.

**Learner:**

There is no need for Learner's existence if all the procedures above always succeed and produce satisfactory results, which is nearly impossible. Learner collects all the failure feedbacks from kernel modules (Syntactic Layer, Semantic Layer and Explainer), and make suggestions on possible **alternative expressions** regarding on different layers.

If unrecognizable characters (either Chinese characters or symbols) appear in the sentence, they are firstly combined and considered as a quotation. Then the sentence is retried by the proceeding procedures, in case that the user forgets or is unwilling to type quotation marks. If this fails again, the characters will be considered as **potential new words** and compared to existing words in keyword-dictionary. Through the comparison, a thesaurus dataset is employed to compute the distance between each two words, and a certain number of words that are nearest to the potential new words are listed as suggestions. The user's selection will cause refreshing of the keyword-dictionary. If this process doesn't work either, the user will be asked to change her/his expression, and the previous one will be regarded as a **special expression** for the same meaning. When special expressions have been accumulated to some amount, the local structural formats will be clustered in semantic layer, so that new rules could be learned and iDian would better accommodate the user's habits. The clustering and learning algorithms are to be implemented in the near future.

Learner also provides an important feature, **online support**. All the databases within iDian can be uploaded and downloaded online, including the keyword-dictionary, the synonym-dictionary, the NUMFORMAT transformation rules, and the general transformation rules. For software developers, they can easily add support for iDian to their new applications by providing **the special suit**, which consists of supplemental dictionaries, specific transformation rules and an Executor that recognizes and deals with actions, primary and secondary objects and conditions sent from Explainer. Special suits can also be shared online, and dynamically integrated into iDian's system, so that skilled users could exchange their knowledge on application operations.

## Discussion

The goal of iDian is to smooth the way to efficient computer operations, and to liberate people from repetitive works in front of the screen. Modern integrated software applications offer much more functions than we actually require, but it's still a difficult task for beginners and intermediate users to find the appropriate method to accomplish their job quickly enough. We are so easily confused in searching a large number of unnecessary buttons and menus for a simple function we assumes to be reasonable, but in most cases we turn to the Internet for help, otherwise we would abandon our ideas or degrade our requirement. With iDian, we can directly input our need and press Enter (just like pressing "*I'm Feeling Lucky*"), and will usually turn out to be "lucky". What iDian provides is a method to handle advanced operation skills with least conditions, far more than a simple search and access tool.

What is more appealing, iDian naturally enables **cross-language operations** of software applications. Since Explainer and Executor are free from specific language, it is possible for an Arabic-speaking user to operate an English application through iDian, even without any knowledge in English

The future development of iDian will continue to focus on improving the language model, which is employed to understand users' command in natural language. Intelligent interaction based on question-and-answer system will be built in, providing a wider range of choices for users to perform their operations. We'll also add up a module that **memorizes and learns** the user's operation and to improve Executor, so that the user could create actions and teach them to iDian. However, it is believed to be a difficult task.